%% file: main.tex
\documentclass{article}

\usepackage{microtype}      
\usepackage{graphicx}
\usepackage{subfigure}
\usepackage{booktabs} 

\usepackage[pagebackref=true,breaklinks=true,colorlinks,bookmarks=false]{hyperref}

\usepackage{times}
\usepackage{epsfig}

\usepackage{amsmath,amsfonts}
\usepackage{amssymb}
\usepackage{amsthm}
\usepackage{nicefrac}       

\usepackage{xcolor}
\usepackage[utf8]{inputenc} 
\usepackage[T1]{fontenc}    
\usepackage{algorithmic}    
\usepackage{algorithm}

\usepackage{enumitem}
\usepackage{commath}
\usepackage{tikz}
\usepackage{thm-restate}

\usetikzlibrary{calc,angles,positioning,intersections,quotes,decorations.markings}
\tikzset{font={\fontsize{14pt}{12}\selectfont}}

\usepackage{pgfplots}
\pgfplotsset{compat=1.16}
\usepackage{arxiv}
\usepackage[square, numbers]{natbib}

\DeclareMathOperator{\sgn}{sgn}
\newcommand{\R}{\mathbb{R}}
\newcommand{\loss}{\mathcal{L}}
\newcommand{\E}{\mathbb{E}}

\newtheorem{assumption}{Assumption}
\usepackage{hyperref}

\begin{document}









\title{Adversarially Robust Learning via \\ Entropic Regularization}
\author{%
  Gauri Jagatap, Ameya Joshi, Animesh Basak Chowdhury, Siddharth Garg, and Chinmay Hegde\thanks{This work was supported in part by NSF grants CCF-2005804 and CCF-1815101.}\\
  New York University\\
  \texttt{\{gauri.jagatap,ameya.joshi,abc586,sg175,chinmay.h\}@nyu.edu} 
}

\maketitle



\input{abstract}
\input{introduction}
\input{priorwork}

\input{problemformulation}
\input{algorithm}
\input{experiments_mnist}

\input{experiments_cifar}
\input{conclusions}

\section{Acknowledgements}
The authors would like to thank Anna Choromanska for useful initial feedback and Praneeth Narayanamurthy for insightful conceptual discussions.
\clearpage

\onecolumn
\appendix
\input{appendix}

\clearpage

\bibliography{biblio}
\bibliographystyle{icml2021}

\end{document}

%% file: abstract.tex
\begin{abstract}
  In this paper we propose a new family of algorithms, ATENT, for training adversarially robust deep neural networks. We formulate a new loss function that is equipped with an additional entropic regularization. Our loss function considers the contribution of adversarial samples that are drawn from a specially designed distribution in the data space that assigns high probability to points with high loss and in the immediate neighborhood of training samples. Our proposed algorithms optimize this loss to seek adversarially robust valleys of the loss landscape. Our approach achieves competitive (or better) performance in terms of robust classification accuracy as compared to several state-of-the-art robust learning approaches on benchmark datasets such as MNIST and CIFAR-10.
\end{abstract}

%% file: introduction.tex
\section{Introduction} \label{sec:intro}

Deep neural networks have led to significant breakthroughs in the fields of computer vision \cite{krizhevsky2012imagenet}, natural language processing \cite{zhang2020adversarial}, speech processing \cite{carlini2016hidden}, recommendation systems \cite{tang2019adversarial} and forensic imaging \cite{rota2016bad}. However, deep networks have also been shown to be very susceptible to carefully designed ``attacks" ~\cite{goodfellow2014explaining,papernot2016transferability,biggio2018wild}. In particular, the outputs of networks trained via traditional approaches are rather brittle to maliciously crafted perturbations in both input data as well as network weights~\cite{biggio2013evasion}. 

Formally put, suppose the forward map between the inputs $x$ and outputs $y$ is modeled via a neural network as $y = f(w;x)$ where $w$ represents the set of trainable weight parameters. 
For a classification task, 
given a labeled dataset $\{x_i, y_i\}$, $i=1,\ldots,n$ where $X$ and $Y$ represents all training data pairs, the standard procedure for training neural networks is to seek the weight parameters $w$ that minimize the empirical risk: 
$$\hat{w} = \arg \min_w \frac{1}{n} \sum_{i=1}^n L(f({w};x_{i}), y_i) := \loss(X;Y,w).$$ 
However, the prediction $\hat{y}(x) = f(\hat{w};x)$ can be very sensitive to changes in both $\hat{w}$ and $x$.  For example, if
a bounded perturbation to a test image input (or to the neural network weights) is permitted, i.e. $\hat{y}_i = f(\hat{w};x_i+\delta_i)$ where $\delta_i$ represents the perturbation, then the predicted label $\hat{y_i}$ can be made \emph{arbitrarily} different from the true label $y_i$.

Several techniques for finding such adversarial perturbations have been put forth. 
Typically, this can be achieved by maximizing the loss function within a neighborhood around the test point $x$~\cite{tramer2017ensemble,madry}: 
\begin{equation}
\bar{x}_{\text{worst}} = \arg\max_{\delta \in \Delta_p} L(f(\hat{w};x+\delta), y)
\label{eq:attack}
\end{equation}
where $\hat{w}$ are the final weights of a pre-trained network. The perturbation set $\Delta_p$ is typically chosen to be an $\ell_p$-ball for some $p \in \{0,1,2,\infty \}$. 

The existence of adversarial attacks motivates the need for a ``defense'' mechanism that makes the network under consideration more robust. Despite a wealth of proposed defense techniques, the jury is still out on how optimal defenses should be constructed~\cite{athalye2018obfuscated}.  

We discuss several families of effective defenses. The first involves \emph{adversarial training}~\cite{madry}. Here, a set of adversarial perturbations of the training data is constructed by solving a min-max objective of the form: 
$$ \hat{w} = \min_w \max_{\delta \in \Delta_p} \frac{1}{n} \sum_{i=1}^n L(f({w};x_{i}+\delta), y_i).$$

\citet{wong2018provable} use a convex outer adversarial polytope as an upper bound for worst-case loss in robust training; here the network is trained by generating adversarial as well as few non-adversarial examples in the convex polytope of the attack via a linear program. Along the same vein include a mixed-integer programming based certified training for piece-wise linear neural networks \cite{tjeng2018evaluating} and integer bound propagation \cite{gowal2019scalable}.

\begin{figure}[!t]
    \centering
\includegraphics[width=\textwidth]{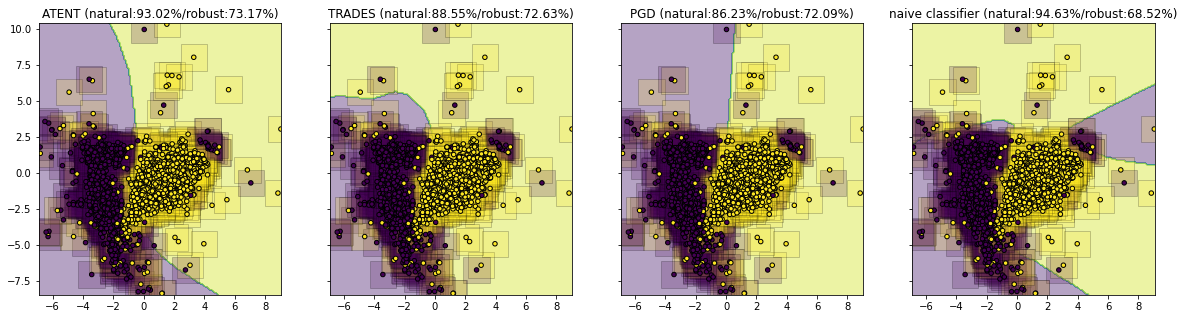}
    \caption{TSNE visualization of decision boundaries for a 3-layer neural network trained using different defenses; corresponding natural and robust test accuracies against $\ell_\infty$ attacks for classifying MNIST digits 5 and 8.}
    \label{fig:my_label}
\end{figure}

The last family of approaches involves \emph{randomized smoothing}. Here, both training the network as well as the inference made by the network are {smoothed} out over several stochastic perturbations of the target example \cite{lecuyer2019certified,kolter,salman2019provably}. This has the effect of optimizing a smoothed-adversarial version of the empirical risk. Randomized smoothing has also been used in combination with adversarial training \cite{bubeck} for improved adversarial robustness under $\ell_2$ attacks\footnote{This family of methods has the additional benefit of being {certifiably robust}: all points within a ball of a given radius around the test point are provably  classified with the correct label.}.


In this paper, we propose a new approach for training adversarially robust neural networks. The key conceptual ingredient underlying our approach is \emph{entropic regularization}. Borrowing intuition from \cite{esgd}, instead of the empirical risk (or its adversarial counterpart), our algorithm instead optimizes over an local entropy-regularized version of the empirical risk:
\begin{align}
    \hat{w} 
    &= \arg \min_w  \mathcal{L}_{DE}, \label{eq:loss} \\
    \mathcal{L}_{DE}& = \int_{X'} \loss(X';Y,w)\left[\frac{e^{\left(\loss(X';Y,w)-\frac{\gamma}{2}\|X-X'\|_p^p\right)}}{Z}\right] dX' .\nonumber
\end{align}
Intuitively, this new loss function can be viewed as the convolution of the empirical risk with a Gibbs-like distribution to sample points from the neighborhoods, $X'$, of the training data points $X$ that have {high} loss. Therefore, compared to adversarial training, we have replaced the inner maximization with an expected value with respect to a modified Gibbs measure which is matched to the {geometry} of the perturbation set.

Since the above loss function is difficult to optimize (or even evaluate exactly), we instead approximate it via Monte Carlo techniques. In particular, we use Stochastic Gradient Langevin Dynamics~\cite{sgld}; in this manner, our approach blends in elements from adversarial training, randomized smoothing, and entropic regularization.  We posit that the combination of these techniques will encourage a classifier to learn a better robust decision boundary as compared to prior art (see visualization in Fig.\ref{fig:my_label}). 

To summarize, our specific contributions are as follows:

\begin{enumerate}[nosep,leftmargin=*]
    \item We propose a new entropy-regularized loss function for training deep neural networks (Eq. \ref{eq:loss}) that is a robust version of the empirical risk.  
     \item We propose a new Monte Carlo algorithm to optimize this new loss function that is based on Stochastic Gradient Langevin Dynamics. We call this approach Adversarial Training with ENTropy (ATENT). 
    \item We show that ATENT-trained networks provide improved (robust) test accuracy when compared to existing defense approaches. 
    \item We combine randomized smoothing with ATENT to show competitive performance with the smoothed version of TRADES.
\end{enumerate}

In particular, we are able to train an $\ell_\infty$-robust CIFAR-10 model to 57.23\% accuracy at PGD attack level $\epsilon = 8/255$, which is higher than the latest benchmark defenses based on both adversarial training using early stopping \cite{bubeck} (56.8\%) as well as TRADES (56.6\%) \cite{trades}. 

%% file: priorwork.tex
\section{Prior Work} \label{sec:priorwork}
Evidence for the existence of adversarial inputs for deep neural networks is by now well established~\cite{Carlini2017cwl2, Dathathri2017minadvexample, adversarialexamples2015,Goodfellow2018existence, Szegedy2014intriguing, MoosaviDezfooli2017UniversalAP}. In image classification, the majority of attacks have focused on the setting where the adversary confounds the classifier by adding an {imperceptible} perturbation to a given input image. The range of the perturbation is pre-specified in terms of bounded pixel-space $\ell_p$-norm balls. Specifically, an $\ell_{p}$- attack model allows the adversary to search over the set of input perturbations $\Delta_{p,\epsilon} = \{\delta : \|\delta\|_p \leq \epsilon\}$ for $p=\{0,1,2,\infty\}$. 

Initial attack methods, including the Fast Gradient Sign Method (FGSM) and its variants~\cite{fgs2014,ifgs2016}, proposed techniques for generating adversarial examples by ascending along the 
sign of the loss gradient:  $$x_{adv} = x + \epsilon\sgn(\nabla_{x} L(f(\hat{w};x), y)),$$ where $(x_{adv}-x) \in \Delta_{\infty,\epsilon}$. Madry \textit{et. al.}~\cite{madry} proposed a stronger adversarial attack via projected gradient descent (PGD) by iterating FGSM several times, such that $$ x^{t+1} = \Pi_{x+\Delta_{p,\epsilon}}(x^{t} +  \alpha\sgn(\nabla_{x} L(f(\hat{w};x), y)),$$ where $p=\{2,\infty\}.$ These attacks are (arguably) the most successful available attack techniques reported to date, and serve as the starting point for our comparisons.
Both Deep Fool \cite{DeepFool} and Carlini-Wagner \cite{carlini2017towards} construct an attack by finding smallest possible perturbation that can flip the label of the network output. 

Several strategies for defending against attacks have been developed. In \cite{madry},  adversarial training is performed via the min-max formulation Eq.~\ref{eq:attack}. The inner maximization is solved using PGD, while the outer objective is minimized using stochastic gradient descent (SGD) with respect to $w$. This can be slow to implement, and speed-ups have been proposed in~\cite{shafahi2019adversarial} and~\cite{wong2020fast}.
In \cite{li2018certified,kolter,lecuyer2019certified,salman2019provably,bubeck}, the authors developed certified defense strategies via randomized smoothing. This approach consists of two stages: the first stage consists of training with noisy samples, and the second stage produces an ensemble-based inference. 
See~\cite{ren2020adversarial} for a more thorough review of the literature on various attack and defense models.

Apart from minimizing the worst case loss, approaches which  minimize the upper bound on worst case loss include \cite{wong2018provable,tjeng2018evaluating,gowal2019scalable}. Another breed of approaches use a modified loss function which considers surrogate adversarial loss as an added regularization, where the surrogate is cross entropy \cite{trades} (TRADES), maximum margin cross entropy \cite{mma} (MMA) and KL divergence \cite{mart} (MART) between adversarial sample predictions and natural sample predictions.  

In a different line of work, there have been efforts towards building neural network networks with improved generalization properties. In particular, heuristic experiments by~\cite{hochreiter1997flat,keskar2016large,li2018visualizing} suggest that the loss surface at the final learned weights for well-generalizing models is relatively ``flat" \footnote{This is not strictly necessary, as demonstrated by good generalization at certain sharp minima \cite{dinh2017sharp}.}. Building on this intuition, Chaudhari \textit{et. al.}~\cite{esgd} showed that by explicitly introducing a smoothing term (via entropic regularization) to the training objective, the learning procedure weights towards regions with flatter minima by design. Their approach, Entropy-SGD (or ESGD), is shown to induce better \emph{generalization} properties in deep networks. We leverage this intuition, but develop a new algorithm for training deep networks with better \emph{adversarial robustness} properties.

%% file: problemformulation.tex

\section{Problem Formulation} \label{sec:probform}

The task of classification, given a training labelled dataset $\{x_i \in \mathcal{X}, y_i \}$, $i\in\{1,\ldots,n\}$, consists of solving the standard objective by optimizing weight parameters $w$, $\min_{w} \frac{1}{n} \sum_{i=1}^n L(f({w};x_{i}), y_i)$ where $y_i$ is a one-hot class encoding vector of length $m$ and $m$ is the total number of classes. The training data matrix itself is represented using shorthand $X \in \R^{n \times d}$ and labels in $Y \in \R^n$ where we have access to $n$ training samples which are $d$-dimensional each. Given this formulation, the primary task is to minimize the cross-entropy Loss function $\loss(w;X,Y)  = -\frac{1}{n}\sum_{i=1}^n \sum_{j=1}^m y_{i,j}\log \hat{y}_{i,j}.$ 
In this paper, we design an augmented version of the loss function $\loss$ which models a class of adversarial perturbations and also introduce a new procedure to minimize it.

%% file: algorithm.tex

We first recap the Entropy SGD \cite{esgd} (see also Appendix \ref{sec:appendix} of the supplement). Entropy-SGD considers an augmented loss function of the form $$\loss_{ent}(w;X,Y) = - \log{\int_{w'} e^{-\loss(w';X,Y)-\frac{\gamma}{2}\|w-w'\|_2^2} dw'}.$$ \vspace{-0.02cm}
By design, minimization of this augmented loss function promotes minima with wide valleys. Such a minimum would be robust to perturbations in $w$, but is not  necessarily advantageous against adversarial data samples $x_{adv}$. In our experiments (Section \ref{sec:exp}) we show that networks trained with Entropy-SGD perform only marginally better against adversarial attacks as compared to those trained with standard SGD.

For the task of adversarial robustness, we instead develop a data-space version of Entropy-SGD. To model for perturbations in the samples, we design an augmented loss that regularizes the {data} space. Note that we only seek specific perturbations of data $x$ that \textit{increase} the overall loss value of prediction. In order to formally motivate our approach, we first make some assumptions. 

\begin{assumption} \label{assump1}
The distribution of possible adversarial data inputs of the neural network obeys a positive exponential distribution of the form below, where the domain of $\loss(X;Y,w)$ is bounded:
\begin{align}  
 &p(X;Y,w,\beta) =\\
  &\begin{cases} \label{eq:prob1}
   Z_{w,\beta}^{-1} e^{\beta \loss(X;Y,w)} & \text{if } \quad \loss(X;Y,w) \leq R ,\\
0          & \text{if } \quad \loss(X;Y,w) > R , 
  \end{cases} \nonumber
\end{align}
and $Z_{w,\beta}$ is the partition function that normalizes the probability distribution.
\end{assumption}
Note here that cross entropy loss $\mathcal{L}$ is always lower bounded as $\loss \geq 0$.

Intuitively, the neural network is more likely to ``see" perturbed examples from the adversary corresponding to higher loss values as compared to lower loss values. The parameter $R$ is chosen to ensure that the integral of the probability curve is bounded. 
When the temperature parameter $\beta \to \infty$, the above Gibbs distribution concentrates at the maximizer(s) of $\loss(\bar{X};Y,w)$, where $\bar{X}$ is the ``worst possible" set of adversarial inputs to the domain of the loss function for fixed weights $w$. For a given attack ball $\Delta_{p,\epsilon}$ with radius $\epsilon$ and norm $p$, and fixed weights $w$, this value equates to:
\begin{align*}
    \bar{X} = \arg \max_{X'\in\Delta_{p,\epsilon}} \mathcal{L}(X';X,Y,w) ,
\end{align*}
where $\max_{X'\in\Delta_{p,\epsilon}} \mathcal{L}(X';X,Y,w)\leq R$.
\begin{assumption} \label{assump:gibbs}
A modified distribution, (without loss of generality, setting $\beta=1$) with an additional smoothing parameter, assumes the form:
\begin{align} 
&p(X';X,Y,w,\gamma) =\\
&
\begin{cases}
Z_{X,w,\gamma}^{-1} e^{\loss(X';Y,w) - \frac{\gamma}{2} \|X'-X\|_F^2}  & \text{if}\quad  \loss(X';Y,w)\leq R \\
0 & \text{if}\quad \loss(X';Y,w)>R
\end{cases} \nonumber
\end{align}
where $Z_{X,w,\gamma}$ is the partition function that normalizes the probability distribution. 
\end{assumption}
Here $\gamma$ controls the penalty of the distance of the adversary from true data $X$; if $\gamma \to \infty$, the sampling is sharp, i.e. $p(X'=X;X,Y,w,\gamma) = 1$ and $p(X'\neq X;X,Y,w,\gamma) = 0$, which is the same as sampling only the standard loss $\mathcal{L}$, meanwhile $\gamma \to 0$ corresponds to a uniform contribution from all possible data points in the loss manifold. 

\begin{figure}
    \centering
    \includegraphics[width=\columnwidth]{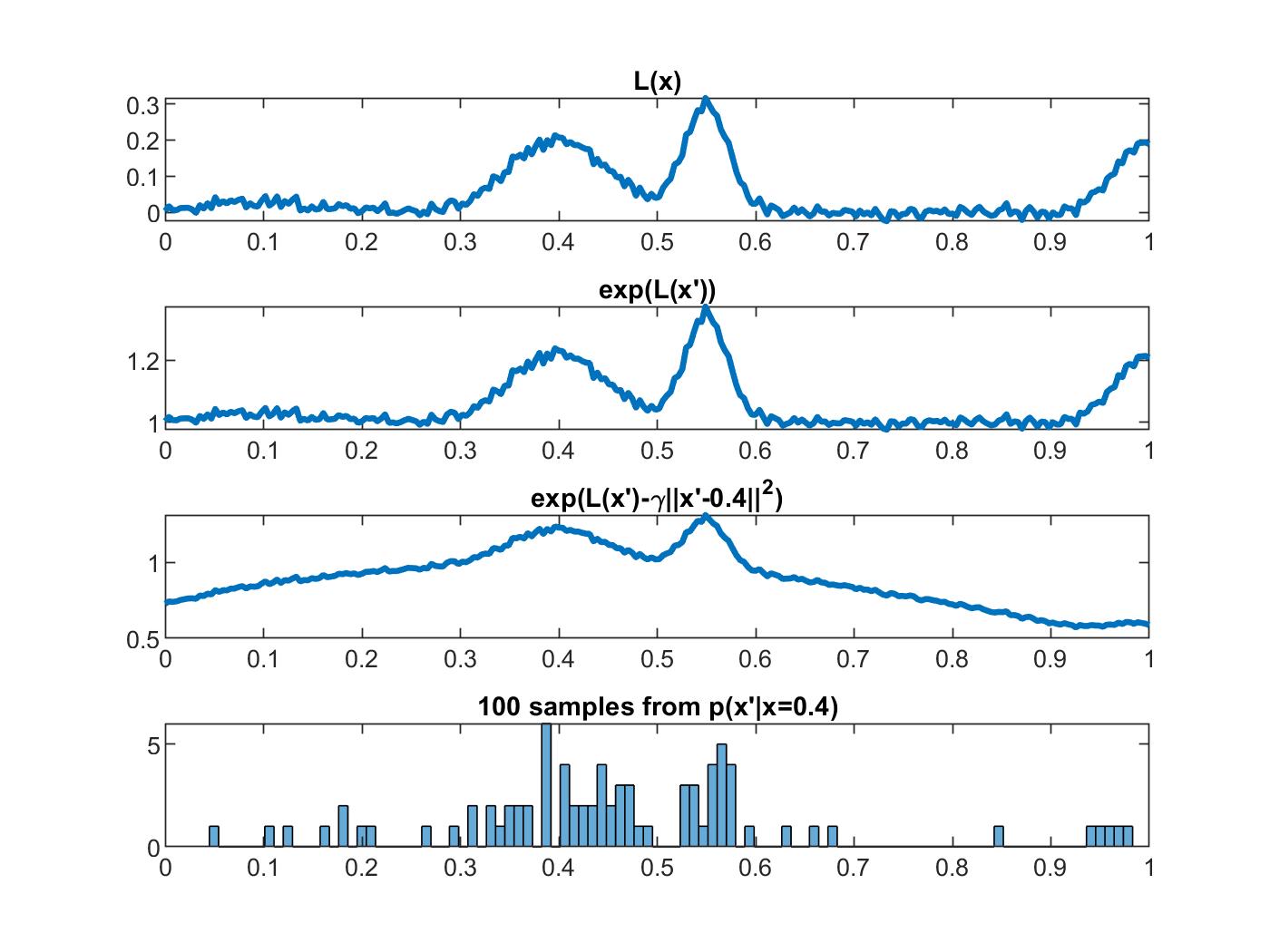}
    \caption{Illustration of the sampling procedure in Assumption \ref{assump:gibbs} at fixed weights $w$. The distribution produces samples $x'$ from distribution $p(x'|x=0.4)$, and we compute the average loss over these samples. Effectively, this encourages ATENT to search for $w$ where $L(x;w)$ is relatively flat in the neighborhood of $x$.}
    \label{fig:sampling}
\end{figure}

Now, we develop an augmented loss function which incorporates the probabilistic formulation in Assumption \ref{assump:gibbs}. The standard objective can be re-written as the functional convolution: 
\begin{align*} 
\min_w \loss(w;X,Y) &:= \min_w \int_{X'} \loss(X';Y,w)\delta(X-X')dX',
\end{align*}
which can be seen as a sharp sampling of the loss function at training 
points $X$. Now, define the \textit{Data-Entropy Loss}:
\begin{align} \label{eq:dataentropyloss}
    \loss_{DE}(w;X,Y,\gamma)    &= \int_{X'} \loss(X';Y,w) p(X';X,Y,w,\gamma) dX'
\end{align}
our new objective is to minimize this augmented objective function $\loss_{DE}(w;X,Y,\gamma)$, which resembles expected value of the standard loss function sampled according to a distribution that (i) penalizes points further away from the true training data (ii) boosts data points which correspond to high loss values. Specifically, the adversarial samples generated by the distribution in Assumption \ref{assump:gibbs} will correspond to those with \textit{high loss} values in the \textit{immediate neighborhood} of the true data samples. This sampling process is also described in Fig. \ref{fig:sampling}. We also highlight theoretical properties of our augmented loss function in Lemma \ref{lem:smooth} of supplement (Appendix \ref{sec:appendix}).

If gradient descent is used to minimize the loss in Eq. \ref{eq:dataentropyloss}, the gradient update corresponding to the augmented loss function can be computed as follows
\begin{align} \nonumber
     &\nabla_w \loss_{de}(w;X,Y,\gamma) \nonumber \\ 
     &= \nabla_w \int_{X'} \loss(X';Y,w) p(X';X,Y,w,\gamma) dX' \nonumber\\ \label{eq:lossgradient}
     &= \nabla_w \mathbb{E}_{X'\sim p(X';X,Y,w,\gamma)} [\loss(X';Y,w)] .
\end{align}


Correspondingly, the weights of the network, when trained using gradient descent, using Eq. \ref{eq:lossgradient}, can be updated as
\begin{align}\label{eq:sgd-update}
    w^+ = w - \eta \nabla_w \mathbb{E}_{X'\sim p(X';X,Y,w,\gamma)} [\loss(X';Y,w)]
\end{align}
where $\eta$ is the step size. The expectation in Eq. \ref{eq:lossgradient} is carried out over the probability distribution of data samples $X'$ as defined in Assumption \ref{assump:gibbs}. This can be seen as the adversarial version of the formulation developed in Entropy SGD \cite{esgd}, where the authors use a Gibbs distribution to model an augmented loss function that explores the loss surface at points that are perturbed from the current weights $w$, denoted by $w'$ (see Appendix \ref{sec:appendix} of the supplement). In contrast, in our approach, we consider loss contributions from perturbations $X'$ of data points $X$. This analogue is driven by the fact that the core objective in \cite{esgd} is to design a network which is robust to perturbations in \textit{weights} (generalization), where as the core objective of this paper is to design a network that is robust to perturbations in the \textit{inputs} (adversarial robustness).

The expectation in Eq. \ref{eq:lossgradient} is computationally intractable to optimize (or evaluate). However, using the Euler discretization of the Langevin Stochastic Differential Equation \cite{sgld}, it can be approximated well. Samples can be generated from $p(X')$ as: 
\begin{align} \label{eq:langevin}
    X'^{k+1} = X'^k + \eta' \nabla_{X'} \log p(X'^t) + \sqrt{2\eta'} \varepsilon \mathcal{N}(0,\mathbb{I})
\end{align}
where $\eta'$ is the step size for Langevin sampling, $\varepsilon$ is a scaling factor that controls additive noise. In Langevin dynamics, when one considers a starting point of $X'^0$ then the procedure above yields samples $X'^1 \dots X'^{t}$ that follow the distribution $p(X')$. Intuitively, the stochastic process $X'^t$ is more likely to visit points in the immediate neighborhoods of the \textit{entire training} dataset $X$  corresponding to high loss values.

Observe that $X'$ and $X$ have the same dimensions and the gradient term in the above equation needs to be computed over $n$, $d$-dimensional data points. In practice this can be computationally expensive. Therefore, we discuss a stochastic variant of this update rule, which considers mini-batches of training data instead.
Plugging in the distribution in Eq.\ref{eq:prob1}, and using the Euler discretization for Langevin Stochastic Differential Equations, the update rule for sampling $X'$ is
\begin{align} \nonumber
    &\frac{X'^{k+1}-X'^k}{\eta'} =\\ \nonumber
    & \nabla_{X'^k} \left(\loss(X'^k;Y,w)-\frac{\gamma}{2}\|X-X'^k\|_F^2\right) + \sqrt{\frac{2\varepsilon^2}{\eta'}} \mathcal{N}(0,\mathbb{I})\\ \label{eq:sampling}
    &= \nabla_{X'^k} \loss(X'^k;Y,w) + \gamma (X-X'^k) + \sqrt{\frac{2\varepsilon^2}{\eta'}} \mathcal{N}(0,\mathbb{I}) 
\end{align}
where we have incorporated $Z_{X,w\gamma}$ in the step size $\eta'$. Note that as the number of updates $k\to \infty$, the estimates from the procedure in Eq. \ref{eq:sampling} converge to samples from the true distribution. $p(X';X,Y,w,\gamma)$. 
We then want to estimate $\nabla_w \loss_{de}(w;X,Y,\gamma) =\nabla_w  \E_{X'\sim p(X')}  \left[\loss(w;X',Y,\gamma)\right]$ using the samples obtained from the above iterative procedure. In \cite{esgd}, authors use an exponentially decaying averaging process to estimate the expected value. 

\textit{\textbf{Batch-wise updates for stochastic gradient estimates:}} As is typical with large datasets, instead of using the entire training data for computing gradients in Eq. \ref{eq:lossgradient} and Eq. \ref{eq:langevin}, one can use batch-wise data where the training data is segmented into $J$ batches $[X_{B_1},X_{B_2}\dots X_{B_J}]$. This is essentially a combination of Stochastic Gradient Descent and Langevin Dynamics and is known as Stochastic Gradient Langevin Dynamics in recent literature \cite{sgld}. 

This discussion effectively leads to the algorithm shown in Algorithm \ref{algo:desgd}, which we refer to as Adversarial Training using Entropy (or ATENT), designed for $\ell_2$ attacks. Note that we have considered exponentially decaying averaging over sample loss $\mu^k$ in Line 10 of Algorithm \ref{algo:desgd}. 

\begin{algorithm}[t]
	\caption{$\ell_2$-ATENT}
	\label{algo:desgd}
	\begin{algorithmic}[1]
		
		\STATE \textbf{Input:} $X = [X_{B_1},X_{B_2}\dots X_{B_J}],f,\eta,\eta', w=w^0, \gamma, \varepsilon, \alpha$
		\FOR{$t=1,\cdots T$} 
        \item[] (outer loop of SGD)
		\FOR{$j=1, \cdots J$}
		\item[] (scan through all batches of data)
		\STATE $x_i^{0} \gets x_i + \delta_{i}$  \COMMENT{$\forall x_i\in X_{B_j}, K$ is number of samples generated using Langevin dynamics}
		\STATE $\mu^j \gets 0$ 
		\FOR{$k = 1, \cdots,K$}
		\STATE	$dx'^k \gets \frac{1}{n_j}\sum_{i=1}^{n_j} \nabla_{x=x'^k}L(f(w^t;x)) +\gamma(x^k-x'^k)$
		\STATE	$x'^{k+1} \gets x'^k + \eta' dx'^k + \sqrt{2\eta'} \varepsilon \mathcal{N}(0,1)$ \COMMENT{Langevin update}
		\STATE $\mu^k \gets \frac{1}{B}\sum_{x_i \in X_{B_j}}  L(w^t;x'^{k+1})$ \COMMENT{augmented batch loss for $X_{B_j}$}
		
		\STATE $\mu^{j} \gets (1-\alpha)\mu^j + \alpha\mu^k$ 
		\ENDFOR 
		\STATE $dL^{t} \gets \nabla_w \mu^{j}$ 
		\STATE $w^{t+1} \gets w^t - \eta dL^t$
		\ENDFOR
		\ENDFOR 
		\STATE \textbf{Output} $\hat{w} \gets w^T$
	\end{algorithmic}
\end{algorithm}

\textit{\textbf{Comparison to PGD Adversarial Training:}} (see also Algorithm \ref{algo:pgd-sgd} \cite{madry} in Appendix \ref{sec:appendixB} (PGD-AT) of the supplement). It is easy to see that the updates of PGD-AT are similar to that of Algorithm \ref{algo:desgd}, consisting broadly of two types of gradient operations in an alternating fashion - (i) an (inner) gradient with respect to samples $X$ (or batch-wise samples $X_{B_j}$) and (ii) an (outer) gradient with respect to weights $w$. While PGD-AT {minimizes} the \textit{worst-case} loss in an $\epsilon$-neighborhood (specifically $\ell_2$ or $\ell_\infty$ ball) of $X$, ATENT {minimizes} an \textit{average loss} over our specifically designed probability distribution (Assumption \ref{assump:gibbs2}) in the neighborhood of $X$. 
Note that the gradient operation in Eq. \ref{eq:sampling} is also the gradient for the regularized version of inner maximation of the adversarial training  problem \cite{madry}, but with added noise term,
\begin{gather} \nonumber
    \max_{X'} \mathcal{L}(X';X,Y,w) \quad s.t. \quad \|X'-X\|_F^2 \leq \epsilon\\
    \Leftrightarrow \max_{X'} \mathcal{L}(X';X,Y,w) - \frac{\gamma}{2}\|X'-X\|_F^2 \label{eq:equi}
\end{gather}
constraint being satisfied if $\|X'-X\|_F$ is minimized, or $-\|X'-X\|_F$ is maximized).

The width of the Gaussian smoothing is adjusted with $\gamma$, which is analogous to controlling the projection radius $\epsilon$ in the inner-maximization of PGD-AT. Then the second and third terms in Eq. \ref{eq:sampling} are simply gradient of an $\ell_2$-regularization term over data space $X'$ and noise. In this way, ATENT can be re-interpreted as a stochastic formalization of $\ell_2$-PGD-AT, with noisy controlled updates.


\textit{\textbf{Comparison to randomized smoothing:}}
In \cite{kolter}, authors describe a defense to adversarial perturbations, in the form of smoothing. A smoothed classifier $g$, under isotropic Gaussian noise $\varepsilon = \mathcal{N}(0,\sigma^2 \mathbb{I})$, produces an output:
\begin{align} \label{eq:smoothing}
    g(x) =  \arg \max_{j} \mathbb{P}(f(x) +\varepsilon) = j).
\end{align}
where $\mathbb{P}$ denotes probability distribution (see Appendix \ref{sec:appendixC} for detailed discussion). SmoothAdv \cite{bubeck} is an adversarial attack as well as defense for smoothed classifiers, which replaces standard loss with cross entropy loss of a smoothed classier. In comparison, we compute a smoothed version of the cross entropy loss of a standard classifier. This is similar to the setup of \cite{blum2020random} (TRADES with smoothing). The procedure in Algorithm\ref{algo:desgd} is therefore  amenable to randomized smoothing in its evaluation. We discuss a smoothed evaluation of ATENT in the next section.
\begin{algorithm}[t]
	\caption{$\ell_\infty$-ATENT}
	\label{algo:atent}
	\begin{algorithmic}[1]
		
		\STATE \textbf{Input:} $X = [X_{B_1},X_{B_2}\dots X_{B_J}],f,\eta,\eta', w=w^0, \gamma, \varepsilon, \alpha$
		\FOR{$t=1,\cdots T$} 
        \item[] (outer loop of SGD)
		\FOR{$j=1, \cdots J$}
		\item[] (scan through all batches of data)
		\STATE $x_i^{0} \gets x_i + \delta_{i}$  \COMMENT{$\forall x_i\in X_{B_j}, K$ is number of samples generated using Langevin dynamics}
		\STATE $\mu^j \gets 0$ 
		\FOR{$k = 1, \cdots,K$}
		\STATE	$dx'^k \gets \frac{1}{n_j}\sum_{i=1}^{n_j} \nabla_{x=x'^k}L(f(w^t;x))$
		\STATE	$x'^{k+1} \gets x'^k + P^K_\gamma(\eta' dx'^k + \sqrt{2\eta'} \varepsilon \mathcal{N}(0,1))$ \COMMENT{update follows Eq.\ref{eq:linfupdate2}, projection active in $K^{th}$ iteration only.}
		
		\STATE $\mu^k \gets \frac{1}{B}\sum_{x_i \in X_{B_j}}  L(w^t;x'^{k+1})$ \COMMENT{augmented batch loss for $X_{B_j}$}
		
		\STATE $\mu^{j} \gets (1-\alpha)\mu^j + \alpha\mu^k$ 
		\ENDFOR 
		\STATE $dL^{t} \gets \nabla_w \mu^{j}$ 
		\STATE $w^{t+1} \gets w^t - \eta dL^t$
		\ENDFOR
		\ENDFOR 
		\STATE \textbf{Output} $\hat{w} \gets w^T$
	\end{algorithmic}
\end{algorithm}

\textit{\textbf{Extension to defense against $\ell_\infty$-attacks:}}
It is evident that due to the isotropic structure of the Gibbs measure around each data point, Algorithm \ref{algo:desgd}, $\ell_2$-ATENT is best suited for $\ell_2$ attacks. However this may not necessarily translate to robustness against $\ell_\infty$ attacks. For this case, one can use an alternate assumption on the distribution of potential adversarial examples. 
\begin{assumption} \label{assump:gibbs2}
We consider a modification of the distribution in Assumption \ref{assump:gibbs} to account for robustness against $\ell_\infty$ type attacks: 
\begin{align} 
&p(X';X,Y,w,\gamma)= \\
& 
\begin{cases}
Z_{X,w,\gamma}^{-1} e^{\left(\loss(X';Y,w) - \frac{\gamma}{2} \|X'-X\|_\infty\right)}  & \text{if}\quad  \loss(X';Y,w)\leq R \\
0 & \text{if}\quad \loss(X';Y,w)>R
\end{cases}\nonumber
\end{align}
where $\|\cdot\|_\infty$ is the $\ell_\infty$ norm on the vectorization of its argument and $Z_{X,w,\gamma}$ normalizes the probability.
\end{assumption}
The corresponding {Data Entropy Loss} for $\ell_\infty$ defenses is:
\begin{align*}
    &\loss_{DE,\infty}(w;X,Y) =\\& Z_{X,w,\gamma}^{-1} \int_{X'}\loss(X';Y,w)e^{\left(\loss(X';Y,w)-\frac{\gamma}{2}\|X-X'\|_\infty\right)} dX'
\end{align*}
This resembles a smoothed version of the loss function with a exponential $\ell_\infty$ kernel along the data dimension to model points in the $\ell_\infty$ neighborhood of $X$ which have high loss. The SGD update to minimize this loss becomes:
\begin{align*}
    \nabla_w \loss_{DE,\infty}(w;X,Y)   &= \nabla_w \E_{X'\sim p(X')} \left[{\loss(w;X',Y)}\right]\\
    \implies w^+ &= w - \eta \nabla_w \loss_{DE,\infty}(w;X,Y)
\end{align*}
where the expectation over $p(X')$ is computed by using samples generated via Langevin Dynamics: 
\begin{align*}
    X'^{k+1} = X'^k + \eta' \nabla_{X'} \log p(X'^k) + \sqrt{2\eta'} \varepsilon\mathcal{N}(0,\mathbb{I})
\end{align*}
Plugging in the distribution in Assumption \ref{assump:gibbs2} the update rule for sampling $X'$:
\begin{align} \nonumber
    &\frac{X'^{k+1} - X'^k}{\eta'} = \\ \nonumber & \nabla_{X'^k} \left(L(X'^k;Y,w)-\frac{\gamma}{2}\|X-X'^k\|_\infty\right) + \sqrt{\frac{2\varepsilon^2}{\eta'}} \mathcal{N}(0,\mathbb{I})\\ \label{eq:linfupdate}
    & \nabla_{X'^k} L(X'^k;Y,w) + \gamma \text{sign}(X_i-X_i'^k)\cdot \mathbf{1} + \sqrt{\frac{2\varepsilon^2}{\eta'}} \mathcal{N}(0,\mathbb{I})
\end{align}
where $i = \arg\max_j |X_j - X_j'^k|$ and $j$ scans all elements of the tensors $X,X'^k$ and $\mathbf{1}_{j} = \delta_{i,j}$. The second term in the update rule navigates the updates $X'^{k+1}$ to lie in the immediate $\ell_\infty$ neighborhood of $X$. Note that this training process requires taking gradients of $\ell_\infty$ distance. In the update rule in Eq. \ref{eq:linfupdate}, the gradient update only happens along one coordinate. In practice when we test this update rule, the algorithm fails to converge. This is due to the fact that typically a sizeable number of elements of $X'-X$ have a large magnitude. 

The expression in the penultimate step of Eq. \ref{eq:linfupdate}, is the gradient of a regularized maximization problem, 
\begin{gather*} \nonumber
\max_{X'} \mathcal{L}(X';X,Y,w) \quad s.t.\quad   \|X'-X\|_\infty \leq \epsilon \\
    \iff \max_{X'} \mathcal{L}(X';X,Y,w) - \gamma \|X'-X\|_\infty 
\end{gather*}
where $\gamma$ is inversely proportional to $\epsilon$ (constraint is satisfied if $\|X'-X\|_\infty$ is minimized, or $-\|X'-X\|_\infty$ is maximized). This expression can be maximized only if $X'\in \Delta_{\infty,\epsilon}$ of $X$; however when we take gradients along only one coordinate, this may not be sufficient to drive all coordinates of $X'$ towards $\Delta_{\infty,\epsilon}$ of $X$. 

Similar to the $\ell_\infty$ Carlini Wagner attack \cite{carlini2017towards}, we replace the gradient update of the $\ell_\infty$ term, with a clipping based projection oracle. We design an accelerated version of the update rule in Eq. \ref{eq:linfupdate}, in which we perform a clipping operation, i.e. an $\ell_\infty$ ball projection of the form:
\begin{align}
 &X'^{k+1} - X'^k \nonumber\\ 
 &= \eta' \nabla_{X'} L(X'^k;Y,w)  + \sqrt{2\eta'} \varepsilon\mathcal{N}(0,\mathbb{I})  , \nonumber\\
 &X'^{K} - X'^{K-1}  \nonumber\\
 &= P{\gamma} \left(\eta' \nabla_{X'} L(X'^{K-1};Y,w)  + \sqrt{2\eta'}\varepsilon\mathcal{N}(0,\mathbb{I}) \right) \label{eq:linfupdate2}
\end{align}
where element-wise projection $P_\gamma(z) = z$ if $|z|<1/\gamma$ and $P_\gamma(z) = 1/\gamma$ if $|z|>1/\gamma$. Empirically, we also explored an alternate implementation where the projection takes place in each inner iteration $k$, however, we find the version in Algorithm \ref{algo:atent} to give better results.

In both Algorithms \ref{algo:desgd} and \ref{algo:atent}, we initialize the Langevin update step with a random normal perturbation $\delta_i$ of benign samples, which is constructed to lie inside within approximately $1/\gamma$ radius of the natural samples.


%% file: experiments_mnist.tex
\section{Experiments}
\label{sec:exp}

In this section we perform experiments on a five-layer convolutional model with 3 CNN and 2 fully connected layers, used in \cite{trades,carlini2017towards}, trained on MNIST. We also train a WideResNet-34-10 on CIFAR10 (as used in \cite{trades}) as well as ResNet20. Due to space constraints, we present supplemental results in Appendix \ref{sec:appendixC}. We conduct our experiments separately on networks specifically trained for $\ell_2$ attacks and those trained for $\ell_\infty$ attacks. We also test randomized smoothing for our $\ell_2$-ATENT model. Source code is provided in the supplementary material.

\textbf{\textit{Attacks:}} For $\ell_2$ attacks, we test PGD-40 with 10 random restarts, and CW2 attacks 
at radius $\epsilon_2=2$ for MNIST and PGD-40 and CW2 attacks at $\epsilon_2=0.43$ ($\approx \epsilon_\infty = 2/255$) and $\epsilon_2=0.5 = 128/255$ for CIFAR10. For $\ell_\infty$ attacks, we test PGD-20, $\ell_\infty$CW, DeepFool attacks at radiii $\epsilon_\infty=0.3$ for MNIST and $\epsilon_\infty=0.031 =  8/255$ for CIFAR10. We test ATENT at other attack radii in Appendix \ref{sec:appendixC}. For implementing the attacks, we use the Foolbox library~\cite{rauber2017foolbox} and the Adversarial Robustness Toolbox~\cite{art2018}.

\textbf{\textit{Defenses:}} We compare models trained using: SGD (vanilla), Entropy SGD \cite{esgd}, PGD-AT \cite{madry} with random starts (or PGD-AT(E) with random start, early stopping \cite{rice2020overfitting}), TRADES \cite{trades}, MMA \cite{mma} and MART \cite{mart}. Wherever available, we use pretrained models to tabulate robust accuracy results for PGD-AT, TRADES, MMA and MART as presented in their published versions. Classifiers giving the best and second best accuracies are highlighted in each category.

\textbf{\textit{Smoothing:}} We also test randomized smoothing \cite{kolter} in addition to our adversarial training to evaluate certified robust accuracies. 

\begin{table}[!t]
    \centering
        \caption{Robust percentage accuracies of 5-layer convolutional net for MNIST against $\ell_2, \epsilon = 2$ attack.}
    \label{tab:MNIST}
    \begin{tabular}{|c|c|c|c|}
    \hline
    Attack$\to$& Benign Acc & $\ell_2$ PGD-40 & $\ell_2$ CW \\
    $\downarrow$ Defense &    &  & \\
    \hline 
         SGD & \textbf{99.38} & 19.40 & 13.20 \\ 
         Entropy SGD & 99.24 & 19.12 & 14.52\\
         $\ell_2$ PGD-AT & 98.76 & 72.94 & -\\
         TRADES & 97.54 & \textbf{76.08} & -\\
         MMA & \textbf{99.27} & 73.02 & 72.72 \\
         \hline
        $\ell_2$ ATENT & 98.66 & \textbf{77.21} & \textbf{76.72} \\
         \hline
    \end{tabular}
\end{table}


\begin{table}[!t]
    \centering
        \caption{Robust accuracies (in percentages) of 5-layer convolutional net for MNIST against $\ell_\infty$, $\epsilon=0.3$ attack.}
    \label{tab:MNIST2}
    \begin{tabular}{|c|c|c|c|}
    \hline
     Attack$\to$ & Benign  & $\ell_\infty$ PGD-20 & $\ell_\infty$ CW \\
    $\downarrow$ Defense & Acc&  $\epsilon_\infty$ = 0.3  & $\epsilon_\infty$ = 0.3  \\
    \hline 
         SGD & 99.39 & 0.97 & 32.37\\ 
         Entropy SGD  & 99.24 & 1.17 & 34.34\\
         $\ell_\infty$ PGD-AT & 99.36 & 96.01 & 94.25\\
         TRADES  & \textbf{99.48} & 96.07 & 94.03\\
         MMA & 98.92 & 95.25 & 94.77 \\
         MART & 98.74 & \textbf{96.48} & \textbf{96.10}\\
        \hline
         $\ell_\infty$ ATENT & \textbf{99.45} & \textbf{96.44} & \textbf{97.40}\\
        \hline
    \end{tabular}
    \vspace{-0.1cm}
\end{table}



The results were generated using an Intel(R) Xeon(R) W-2195 CPU 2.30GHz Lambda cluster with 18 cores and a NVIDIA TITAN GPU running PyTorch version 1.4.0.

\subsection{MNIST}

In Tables \ref{tab:MNIST} and \ref{tab:MNIST2}, we tabulate the robust accuracy for 5-layer convolutional network trained using the various approaches discussed above for both $\ell_2$ and $\ell_\infty$ attacks respectively. 

\textit{\textbf{Training setup:}} Complete details are provided in Appendix \ref{sec:appendixC}. Our experiments for $\ell_2$ attack are presented in Table \ref{tab:MNIST}. We perform these experiments on a LeNet5 model imported from the Advertorch toolbox (architecture details are provided in the supplement). For $\ell_2$-ATENT we use a batch-size of 50 and SGD with learning rate of $\eta= 0.001$ for updating weights.  We set $\gamma=0.05$ and noise $\varepsilon \sim 0.001\mathcal{N}(0,\mathbb{I})$. We perform $K=40$ Langevin epochs and set the Langevin parameter $\alpha=0.9$, and step $\eta'=0.25$. For attack, we do a 40-step PGD attack with $\ell_2$-ball radius of $\epsilon = 2$. The step size for the PGD attack is 0.25, consistent with the configuration in \cite{mma}. We perform early stopping by tracking robust accuracies of validation set and report the best accuracy found.    

In Table \ref{tab:MNIST2}, we use a SmallCNN configuration as described in \cite{trades} (architecture in supplement). We use a batch-size of 128, SGD optimizer with learning rate of $\eta= 0.01$ for updating weights. We set $\gamma=3.33$ and noise $\varepsilon \sim 0.001\mathcal{N}(0,\mathbb{I})$. We perform $L=40$ Langevin epochs and we set the Langevin parameter $\alpha=0.9$, and step $\eta'=0.01$, consistent with the configuration in \cite{trades}. For the PGD attack, we use a 20-step PGD attack with step-size $0.01$, for $\ell_\infty$-ball radius of $\epsilon = 0.3$. We perform an early stopping by tracking robust accuracies on the validation set and report the best accuracy found. Other attack configurations can be found in the supplement.   

Our experiments on the Entropy-SGD (row 2 in Tables \ref{tab:MNIST} and \ref{tab:MNIST2}) trained network suggests that networks trained to find flat minima (with respect to weights) are not more robust to adversarial samples as compared to vanilla SGD. 

%% file: experiments_cifar.tex
\subsection{CIFAR10}
Next, we extend our experiments to CIFAR-10 using a WideResNet 34-10 as described in \cite{trades,mart} as well as ResNet-20. For PGD-AT (and PGD-AT (E)), TRADES, and MART, we use the default values stated in their corresponding papers. 

\textbf{\textit{Training setup:}} Complete details in Appendix \ref{sec:appendixC}. 
Robust accuracies of WRN-34-10 classifer trained using state of art defense models are evaluated at the $\ell_\infty$ attack benchmark requirement of radius $\epsilon = 8/255$, on CIFAR10 dataset and tabulated in Table \ref{tab:CIFAR2}. For $\ell_\infty$-ATENT, we use a batch-size of 128, SGD optimizer for weights, with learning rate $\eta = 0.1$ (decayed to 0.01 at epoch 76), 76 total epochs, weight decay of $5\times 10^{-4}$ and momentum 0.9. We set $\gamma = 1/(0.0031)$, $K=10$ Langevin iterations, $\varepsilon = 0.001\mathcal{N}(0,\mathbb{I})$, at step size $\eta' = 0.007$. We test against 20-step PGD attack, with step size 0.003, as well as $\ell_\infty$-CW and Deep Fool attacks using FoolBox. $\ell_\infty$-ATENT is consistently among the top two performers at benchmark configurations.

\textit{\textbf{Importance of early stopping:}} Because WRN34-10 is highly overparameterized with approximately 48 million trainable parameters, it tends to overfit adversarially-perturbed CIFAR10 examples. The success of TRADES (and also PGD) in \cite{rice2020overfitting} relies on an early stopping condition and corresponding learning rate scheduler. We strategically search different early stopping points and report the best possible robust accuracy from different stopping points.

We test efficiency of our $\ell_2$-based defense on both $\ell_2$ attacks, as well as compute $\ell_2$ certified robustness for the smoothed version of ATENT against smoothed TRADES \cite{blum2020random} in Table \ref{tab:smooth} in Appendix \ref{sec:appendixC}. We find that our formulation of $\ell_2$ ATENT is both robust against $\ell_2$ attacks, as well as gives a competitive certificate against adversarial perturbations for ResNet20 on CIFAR10.



\begin{table}[!t]
    \centering
        \caption{Robust accuracies of WRN34-10 net for CIFAR10 against $\ell_\infty$ attack of $\epsilon=8/255$.}
    \label{tab:CIFAR2}
    \begin{tabular}{|c|c|p{0.07\textwidth}|p{0.05\textwidth}||p{0.06\textwidth}|}
    \hline
    Defense$\to$  & PGD & TRADES  & MART & ATENT\\
    $\downarrow$ Attack  & AT & & & $\ell_\infty$\\
    \hline 
    \hline
         Benign & \textbf{87.30}  & 84.92 &  84.17 & \textbf{85.67} \\ 
         \hline
         $\ell_\infty$ PGD-20 & 47.04 & 56.61 & \textbf{57.39} & \textbf{57.23} \\
         (E) & 56.80 & & & \\
         \hline
         $\ell_\infty$ CW  & 49.27 & \textbf{62.67}   & 54.53 & \textbf{62.34} \\
         \hline 
         $\ell_\infty$ DeepFool  & - & \textbf{58.15}  & 55.89 & \textbf{57.21}\\
        \hline
    \end{tabular}

\end{table}

In Appendix \ref{sec:appendixC} we also demonstrate a fine-tuning approach for ATENT, where we consider a pre-trained WRN34-10 and fine tune it using ATENT, similar to the approach in \cite{jeddi2020simple}. We find that ATENT can be used to fine tune a naturally pretrained model at lower computational complexity to give competitive robust accuracies while almost retaining the performance on benign data.




%% file: conclusions.tex

\subsection{Discussion}
We propose a new algorithm for defending neural networks against adversarial attacks. We demonstrate competitive (and often improved) performance of our family of algorithms (ATENT) against the state of the art. We analyze the connections of ATENT with both PGD-adversarial training as well as randomized smoothing. Future work includes extending to larger datasets such as ImageNet, as well as theoretical analysis for algorithm convergence.

%% file: appendix.tex
\section{Additional experiments and details} \label{sec:appendixC}

In this section, we provide additional details as well as experiments to supplement those in Section \ref{sec:exp}.

\begin{table}[!t]
    \centering
        \caption{Percentage robust accuracies of ResNet-20 for CIFAR10 against $\ell_2$ attack}
    \label{tab:CIFAR1}
    \begin{tabular}{|c|c|c|c|c|c|}
    \hline
    $\downarrow$ Algorithm/Attack$\to$ & Model & Training param & Benign & PGD-10  & PGD-10 \\
    & & & & $\epsilon_2$ = 0.5 & $\epsilon_2$ = 1 \\
    \hline 
         PGD-AT & WideResNet28-4 & $\epsilon_2 = 1$ & 83.25 & 66.69 & 46.11\\
         MMA & WideResNet28-4 & $d = 1$ & 88.92 & 66.81 & 37.22\\
         \hline 
         $\ell_2$ ATENT & ResNet20 & $\gamma = 0.05$, $\varepsilon=0.001\mathcal{N}(0,\mathbf{1})$ & 85.44 & 65.12 & 47.38 \\         
         \hline
         &  &  &  & & $\epsilon_2$=0.435\\
         \hline
         TRADES (smooth) & ResNet20 & $\epsilon_2 = 0.435$, $\sigma=0.12$ & 75.13 & & 61.03 \\
         \hline
         $\ell_2$ ATENT & ResNet20 & $\gamma = 0.05$, $\sqrt{2\eta'}\varepsilon=0.12\mathcal{N}(0,\mathbf{1})$ & 72.10 & & 64.53\\
         \hline
    \end{tabular}

\end{table}

\subsection{Detailed training setup}
\textbf{\textit{Architectures:}} For MNIST- $\ell_\infty$ experiments, we consider a CNN architecture with the following configuration (same as \cite{trades}).  Feature extraction consists of the following sequence of operations: two layers of 2-D convolutions with 32 channels, kernal size 3, RelU activation each, followed by maxpooling by factor 2, followed by two layers of 2-D convolutions with 64 channels, kernel size 3, ReLU activation, and finally another maxpool (by 2) operation. This is followed by the classification module, consisting of a fully connected layer of size 1024 $\times$ 200, ReLU activation, dropout, another fully connected layer of size $200 \times 200$, ReLU activation and a final fully connected layer of size $200 \times 10$. Effectively this network has 4 convolutional and 3 fully connected layers. We use batch size of 128 with this configuration.

For MNIST-$\ell_2$ experiments, we consider the LeNet5 model from the Advertorch library (same as \cite{mma}). This consists of a feature extractor of the form - two layers of 2-D convolutions, first one with 32 and second one with 64 channels, ReLU activation and maxpool by factor 2. The classifier consists of one fully connected layer of dimension $3136 \times 1024$ followed by ReLU activation, and finally another fully connected layer of size $1024\times 10$. We use batch size of 50 with this configuration.

For CIFAR-$\ell_\infty$ experiments we consider a WideResNet with 34 layers and widening factor 10 (same as \cite{trades} and \cite{madry}). It consists of a 2-D convolutional operation, followed by 3 building blocks of WideResNet, ReLU, 2D average pooling and fully connected layer. Each building block of the WideResNet consists of 5 successive operations of batch normalization, ReLU, 2D convolution, another batch normalization, ReLU, dropout, a 2-D convolution and shortcut connection. We use batch size of 128 with this configuration.

For CIFAR-$\ell_2$ experiments, we consider a ResNet with 20 layers. This ResNet consists of a 2-D convolution, followed by three blocks, each consisting of 3 basic blocks with 2 convolutional layers, batch normalization and ReLU. This is finally followed by average pooling and a fully connected layer. We use batch size of 256 with this configuration. 

\textbf{\textit{Training SGD and Entropy SGD models for MNIST experiments:}} For SGD, we trained the 7-layer convolutional network setup in \cite{trades,carlini2017towards} with the MNIST dataset, setting batch size of 128, for $\ell_\infty$ SGD optimizer using a learning rate of 0.1, for 50 epochs. For Entropy SGD, with 5 langevin steps, and $\gamma=10^{-3}$, batch size of 128 and learning rate of 0.1 and 50 total epochs.

\subsection{$\ell_2$ ATENT}
\textit{\textbf{$\ell_2$-PGD attacks on CIFAR10:}} We explore the effectiveness of $\ell_2$-ATENT as a defense against $\ell_2$ perturbations. These results are tabulated in Table \ref{tab:CIFAR1}. We test 10-step PGD adversarial attacks at $\epsilon_2= 0.5$ and $\epsilon_2 = 1$. For the purpose of this comparison, we compare pretrained models of MMA and PGD-AT at $\epsilon_2 = 1$. To train ATENT, we use $\gamma=0.08$ for $\epsilon_2=1$ 10 step attack (with $2.5\epsilon_2/10$ step size), $K=10$ langevin iterations, langevin step $\eta'=2 \epsilon_2/K$, learning rate for weights $\eta = 0.1$. We also compare models primarily trained to boost the certificate of randomized smoothing. For this we train a ResNet20 model for both TRADES (at default parameter setting) and $\ell_2$ ATENT, at $\eta'=$ and $\gamma=$, such that the effective noise standard deviation is $0.12$. These models are tested against PGD-10 attacks at radius $\epsilon_2= 0.435$. In all $\ell_2$ ATENT experiments, we choose the value of $\gamma$ such that the perturbation $\|X'^{K}-X\|_F \approx \epsilon_2$ of corresponding models of TRADES and PGD-AT. For all ATENT experiments, we set $\alpha=0.9$.

\textbf{\textit{Experiments on randomized smoothing:}} Since the formulation of ATENT is similar to a noisy PGD adversarial training algorithm, we test its efficiency towards randomized smoothing and producing a higher robustness certificate (Table \ref{tab:smooth}). For this we train a ResNet-20 on CIFAR10, at $\gamma=0.05, \eta'=0.02, \eta=0.1, K=10$, and tune the noise $\varepsilon$, such that effective noise $\sqrt{2\eta'}\varepsilon$ has standard deviation $\sigma=0.12$. We compare the results of randomized smoothing to established benchmarks on ResNet-110 (results have been borrowed from Table 1 of \cite{blum2020random}). as well as a smaller ResNet-20 model trained using TRADES at its default settings. We observe that without any modification to the current form of ATENT, our method is capable of producing a competitive certificate to state of art methods. In future work we aim to design modifications to ATENT which can serve the objective of certification. 

\begin{table}[!t]
    \centering
        \caption{Smoothed robust accuracies for CIFAR10 against $\ell_\infty$ attack of $\epsilon=2/255$ ($\ell_2, \epsilon=0.435$), smoothing factor $\sigma = 0.12$.}
    \label{tab:smooth}
    \begin{tabular}{|c|c|c|c|c|}
    \hline
    Smoothing radius $\to$ & ResNet & Standard & $\epsilon_\infty $\\
    $\downarrow$ Defense & Type & 0 & 2/255\\
    \hline 
    Crown IBP \cite{zhang2019towards} & 110 & 72.0 & 54.0 \\
    Smoothing \cite{wong2018scaling} & 110 & 68.3 & 53.9\\
    SmoothAdv \cite{salman2019provably} & 110 & \textbf{82.1} & \textbf{60.8} \\
    TRADES Smoothing \cite{blum2020random}& 110 & \textbf{78.7} & \textbf{62.6}\\
    \hline 
    \hline 
    TRADES Smoothing & 20 &  \textbf{78.2} & \textbf{58.1} \\
    \hline
    ATENT (ours) & 20 & \textbf{72.2}  & \textbf{55.41}\\
    \hline
    \end{tabular}

\end{table}

\subsection{$\ell_\infty$ ATENT}

\begin{figure}[!h]
    \centering
    \includegraphics[width=0.4\textwidth]{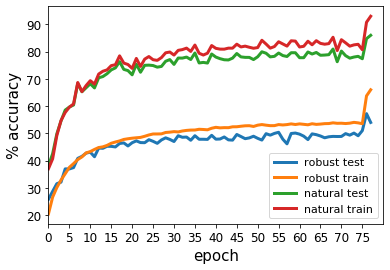}
    \caption{Benign training, test and robust training. test accuracies of ATENT. The learning rate is decayed at epoch 76, where the robust test accuracy peaks. This is the accuracy reported.}
    \label{fig:training}
\end{figure}

\textit{\textbf{Training characteristics of $\ell_\infty$ ATENT:}} In Figure \ref{fig:training} we display the training curves of ATENT. As shown, the robust accuracies spike sharply after the first learning rate decay, followed by an immediate decrease in robust accuracies. This behavior is similar to that observed in \cite{rice2020overfitting}. This is also the key intuition used in the design of the learning rate scheduler for TRADES.

\begin{table}[!b]
    \centering
        \caption{Percentage robust accuracies for CIFAR10 against $\ell_\infty$ PGD and ATENT attacks of different radii.}
    \label{tab:atent}
    \begin{tabular}{|c|c|c|c|c|c|}
    \hline
    Attack radius $\to$ & 2/255 & 4/255 & 8/255 & 12/255\\
    $\downarrow$ Attack & & & &\\
    \hline 
    PGD-20 & 79.83 & 73.35 & 57.23 & 39.37 \\
    ATENT-20 & 79.95 & 73.76 & 59.69  & 47.53\\
    \hline
    \end{tabular}

\end{table}
\textit{\textbf{ATENT as Attack:}} For our $\ell_\infty$-ATENT WideResNet-34-10, we also test $\ell_\infty$-ATENT as an attack. We keep the same configuration as that of PGD-20, for ATENT. We compare the performance of our $\ell_\infty$-ATENT trained model (specifically designed to work against $\epsilon_\infty$=8/255 attacks). The values (Table \ref{tab:atent}) suggest that the adversarial perturbations generated by ATENT are similar in strength to those produced by PGD (worst possible attack).

\textit{\textbf{Computational complexity:}}
In terms of computational complexity, ATENT matches that of PGD and TRADES, as can be observed from the fact that all three approaches are nested iterative optimizations. Due to the high computational complexity of all adversarial algorithms, we test a fine-tuning approach, to trade computational complexity for accuracy. This method is suggested in \cite{jeddi2020simple}. In this context, we take a pre-trained WideResNet-34-10 which has been trained on benign CIFAR10 samples only. This model is then fine tuned on adversarial training data, via $\ell_\infty$ ATENT using a low learning rate $\eta=0.0001$ and trained for only 20 epochs. The final robust accuracy at $\epsilon_\infty = 8/255$  is  $52.1\%$. This is accuracy marginally improves upon the robust accuracy observed (51.7\%) for fine-tuned WideResNet-28-10 PGD-AT trained model in \cite{jeddi2020simple}. This experiments suggests that ATENT is amenable for fine tuning pretrained benign models using lesser computation, but at the cost of slightly reduced robust accuracy (roughly $5\%$ drop at benchmark of $\epsilon_\infty = 8/255$).


\section{Proofs and Derivations} \label{sec:appendix}

\subsection{Theoretical properties of the augmented loss}

We now state an informal theorem on the conditions required for convergence of SGLD in Eq. \ref{eq:langevin} for estimating adversarial samples $X'$.

\begin{restatable}{lem}{smooth} \label{lem:smooth}
The effective loss $F(X';X,Y,w) := \frac{\gamma}{2}\|X-X'\|_F^2 - \mathcal{L}(X';Y,w)$ which guides the Langevin sampling process in Eq. \ref{eq:sampling} is
\begin{enumerate}
    \item $\beta+\gamma$ smooth if $\mathcal{L}(X;Y,w)$ is $\beta$-smooth in $X$.
    \item $\left(\frac{\gamma}{4},\frac{L^2}{\gamma} + \frac{\gamma}{2}\|X\|_F^2\right)$ dissipative if $\mathcal{L}(X;Y,w)$ is $L$-Lipschitz in $X$.
\end{enumerate}
\end{restatable}
One can then use smoothness and dissipativity of $F(X';Y,w)$ to show convergence of SGLD for the optimization over $X'$ (Eq. \ref{eq:sampling}) via Theorem 3.3 of \cite{xu2017global}. 

We first derive smoothness conditions for the effective loss 
$$F(X';X,Y,w) := \frac{\gamma}{2}\|X-X'\|_F^2 - \mathcal{L} (X';Y,w), \quad \forall X'_1,X_2'.$$ 
We use abbreviations $p(X') := p(X';X,Y,w), F(X') := F(X';X,Y,w), \mathcal{L}(X';Y,z) := \mathcal{L}(X')$ and $\mathcal{L}(X;Y,z) := \mathcal{L}(X)$, and assume that $X$ and $X'$ are vectorized. Unless specified otherwise, $\|\cdot\|$ refers to the vector 2-norm.
\begin{proof}
Let us show that $\|\nabla_{X'}F(X'_2) - \nabla_{X'}F(X'_1)\| \leq \beta'\|X_2'-X_1'\|$. If the original loss function is $\beta$ smooth, i.e.,
\begin{align*}
\|\nabla_{X'}\mathcal{L}(X_2')-\nabla_{X'}\mathcal{L}(X_2')\| &\leq \beta \|X_2'-X_1'\| ,
\end{align*}
then: 
\begin{align*}
    \|\nabla_{X'}F(X'_2) - \nabla_{X'}F(X'_1)\| &\leq \|-\nabla_{X'} \mathcal{L}(X'_2) + \nabla_{X'}\mathcal{L}(X'_1) - \gamma(X - X'_2) + \gamma(X-X_1')\| \\
    &\leq \|\nabla_{X'} \mathcal{L}(X'_2) - \nabla_{X'}\mathcal{L}(X'_1)\| + \| \gamma( X'_2-X_1')\| \\
    &\leq (\beta+\gamma) \|X_2'-X_1'\|
\end{align*}
by application of the triangle inequality.

Next, we establish conditions required to show $(m,b)$-dissipativity for $F(X')$, i.e. $\langle\nabla_{X'}F(X'),X'\rangle \geq m\|X'\|^2_2 - b$ for positive constants $m,b>0$, $\forall X'$. 
To show that:
\begin{align*}
    \langle\nabla_{X'}F(X'),X'\rangle &\geq m\|X'\|^2_2 - b 
\end{align*}
where the left side of inequality can be expanded as:
\begin{align} \nonumber
\langle\nabla_{X'}F(X'),X'\rangle
    &= \langle-\nabla_{X'}\loss(X') + \gamma(X'-X),X'\rangle \\ \nonumber
    &=\langle-\nabla_{X'}\loss(X'),X'\rangle + \gamma\|X'\|_2^2 -  \gamma \langle X,X'\rangle
\\ \nonumber    &= \langle-\nabla_{X'}\loss(X'),X'\rangle + \gamma\|X'\|_2^2 -  \frac{\gamma}{2} \left(\|X'\|_2^2 + \|X\|_2^2- \|X-X'\|_2^2\right)\\ \nonumber
    &\geq \langle-\nabla_{X'}\loss(X'),X'\rangle + \gamma\|X'\|_2^2 -  \frac{\gamma}{2} \left(\|X'\|_2^2 + \|X\|_2^2\right)\\ \label{eq:ineq1}
    &=\langle-\nabla_{X'}\loss(X'),X'\rangle + \frac{\gamma}{2}\|X'\|_2^2 -  \frac{\gamma}{2} \|X\|_2^2
    \end{align}
To find the inner product $\langle-\nabla_{X'}\loss(X'),X'\rangle$, we expand squares:
\begin{align*}
      \|\nabla_{X'}\loss(X') - \frac{\gamma}{2}X'\|^2 = \|\nabla_{X'}\loss(X')\|_2^2 + \frac{\gamma^2}{4}\|X'\|_2^2 - \gamma \langle\nabla_{X'}\loss(X'),X'\rangle &\geq 0\\
      -\langle\nabla_{X'}\loss(X'),X'\rangle &\geq -\frac{\|\nabla_{X'}\loss(X')\|_2^2}{\gamma} - \frac{\gamma}{4}\|X'\|_2^2
\end{align*}
Plugging this into \eqref{eq:ineq1}, and assuming Lipschitz continuity of original loss $\mathcal{L}(X')$, i.e., $\|\nabla_{X'}\mathcal{L}(X')\|_2 \leq L$:
\begin{align*}
\langle\nabla_{X'}F(X'),X'\rangle  &\geq
    \langle-\nabla_{X'}\loss(X'),X'\rangle + \frac{\gamma}{2}\|X'\|_2^2 -  \frac{\gamma}{2} \|X\|_2^2 \\
    &\geq -\frac{\|\nabla_{X'}\loss(X')\|_2^2}{\gamma} - \frac{\gamma}{4}\|X'\|_2^2 + \frac{\gamma}{2}\|X'\|_2^2 -  \frac{\gamma}{2} \|X\|_2^2 \\
    &= \frac{\gamma}{4} \|X'\|_2^2-\left(\frac{L^2}{\gamma} +\frac{\gamma}{2}\|X\|_2^2\right) \\
    &=m\|X'\|_2^2 - b
\end{align*}
where $m=\frac{\gamma}{4}$ and $b = \frac{L^2}{\gamma} + \frac{\gamma}{2}\|X\|_2^2$. Thus, $F(X')$ is $(\frac{\gamma}{4},\frac{L^2}{\gamma} + \frac{\gamma}{2}\|X\|_2^2)$ dissipative, if $\loss(X')$ is $L$-Lipschitz.

\end{proof}

With Lemma \ref{lem:smooth} we can show convergence of the SGLD inner optimization loop. To minimize overall loss function, the data entropy loss $\loss_{DE}$ is minimized w.r.t. $w$, via Stochastic Gradient Descent (SGD). The gradient update for weights $w$ are designed via \eqref{eq:lossgradient} as follows: 
\begin{align*} 
     \nabla_w \loss_{DE}(w;X,Y,\gamma) 
     &= \nabla_w \int_{X'} \loss(X';Y,w) p(X';X,Y,w,\gamma) dX' = \nabla_w \mathbb{E}_{X'\sim p(X';X,Y,w,\gamma)} [\loss(X';Y,w)] \\
     &=  \int_{X'} \nabla_w\left( \loss(X';Y,w) p(X';X,Y,w,\gamma) \right) dX' \\
     &=  \int_{X'} \nabla_w \loss(X';Y,w) \cdot  p(X';X,Y,w,\gamma)  + \nabla_w \loss(X';Y,w) \cdot \loss(X';Y,w) \cdot p(X';X,Y,w,\gamma) dX'\\
     &=\int_{X'} \nabla_w \loss(X';Y,w)  \cdot \left(\loss(X';Y,w)+1\right) \cdot p(X';X,Y,w,\gamma) dX' \\
     &=   \mathbb{E}_{X'\sim p(X';X,Y,w,\gamma)} \left( \nabla_w\loss(X';Y,w)  \cdot (\loss(X';Y,w)+1)\right) 
\end{align*}
Then a loose upper bound on Lipschitz continuity of $\loss_{DE}$ is $\| \nabla_w \loss_{DE}(w;X,Y,\gamma) \|_2 \leq \bar{L}(R+1)$, if original loss is $\bar{L}$-Lipschitz in $w$ and $\loss(X) \leq R$. Due to the complicated form of this expression, establishing $\beta$-smoothness will require extra rigor. We push a more thorough evaluation of the convergence of the outer SGD loop to future work.

\subsection{Entropy SGD}

\begin{algorithm}[th]
	\caption{Entropy SGD}
	\label{algo:esgd}
	\begin{algorithmic}[1]
		
		\STATE \textbf{Input:} $X = [X_{B_1},X_{B_2}\dots X_{B_J}],f,\eta,\eta', w=w^0, \gamma, \alpha, \varepsilon$
		\FOR{$t=0,\cdots T-1$}
		\FOR{$j=1, \cdots J$}
		\STATE $w'^{0} \gets w^t, \mu^0 \gets w^t$  \hspace{1cm}  \COMMENT{Repeat inner loop for all training batches $j$}
		\FOR{$k = 0, \cdots,K-1$}
		\STATE	$dw'^k \gets \frac{1}{n_j}\sum_{i=1}^{n_j} -\nabla_{w=w^k}L(f(w;x_i)) + \gamma(w^k-w'^k)$  \COMMENT{$\forall x_i \in X_{B_j}$}
		
		\STATE	$w'^{k+1} \gets w'^k + \eta' dw'^k + \sqrt{2\eta'} \varepsilon \mathcal{N}(0,1)$ \COMMENT{Langevin update}
		
		\STATE $\mu^k \gets (1-\alpha)\mu^k + \alpha w'^{k+1}$
		\ENDFOR
		\STATE $\mu^t \gets \mu^K$
		\STATE $w^{t+1} \gets w^t - \eta \gamma (w^t - \mu^t)$ \COMMENT{Repeat outer loop step for all training batches $j$}
		\ENDFOR
		\ENDFOR 
		\STATE \textbf{Output} $\hat{w} \gets w^T$
	\end{algorithmic}
\end{algorithm}

In \cite{esgd} authors claim that neural networks that favor wide local minima have better generalization properties, in terms of perturbations to data, weights as well as activations. Mathematically, the formulation in Entropy SGD can be summarized as follows. A basic way to model the distribution of the weights of the neural network is using a Gibbs distribution of the form:
\begin{align*}
    p(w;X,Y,\beta) = Z_{X,\beta}^{-1} \exp{\left(-\beta \loss(w;X,Y)\right)}
\end{align*}
When $\beta \to \infty$, this distribution concentrates at the global (if unique) minimizer of $\loss(w^*;X,Y)$. A modified Gibbs distribution, with an additional smoothing parameter is introduced, which assumes the form: 
\begin{align}
    p(w';w,X,Y,\beta=1,\gamma) = Z_{w,X,\gamma}^{-1} \exp{\left(- \loss(w';X,Y) - \frac{\gamma}{2} \|w'-w\|_2^2\right)}
\end{align}
where $Z_{w,X,\gamma}$ normalizes the probability.

Here $\gamma$ controls the width of the valley; if $\gamma \to \infty$, the sampling is sharp, and this corresponds to no smoothing effect, meanwhile $\gamma \to 0$ corresponds to a uniform contribution from all points in the loss manifold. The standard objective is:
\begin{align*}
\min_w \loss(w;X,Y) &:= \min_w - \log{\left(\exp{\left(-\loss(w;X,Y)\right)}\right)}\\
&= \min_w - \log{\left(\int_{w'} \exp{\left(-\loss(w';X,Y)\right)}\delta(w-w')dw'\right)} 
\end{align*}
which can be seen as a sharp sampling of the loss function. Now, if one defined the Local Entropy as:
\begin{align*}
    \loss_{ent}(w;X,Y) &= -\log (Z_{w,X,Y,\gamma}) \\
    &= - \log{\left(\int_{w'} \exp{\left(-\loss(w';X,Y)-\frac{\gamma}{2}\|w-w'\|_2^2\right)} dw'\right)}
\end{align*}
our new objective is to minimize this augmented objective function $\loss_{ent}(w;X,Y)$, which resembles a smoothed version of the loss function with a Gaussian kernel. The SGD update can be designed as follows:
\begin{align*}
    \nabla_w \loss_{ent}(w;X,Y) &= -\nabla_w (\log (Z_{w,X,Y,\gamma}))\\
    &= Z_{w,X,\gamma}^{-1} \nabla_w(Z_{w,X,\gamma}) \\
    &= Z_{w,X,\gamma}^{-1} \left(\int_{w'} \exp{\left(-\loss(w';X,Y)-\frac{\gamma}{2}\|w-w'\|_2^2\right)} \cdot \gamma (w-w') dw'\right)\\
    &= \int_{w'} p(w';w,X,Y,\gamma) \cdot \gamma (w-w') dw' \\
    &= \E_{w'\sim p(w')} \left[\gamma (w-w')\right]
\end{align*}
Then, using this gradient, the SGD update for a given batch is designed as:
\begin{align*}
    w^+ = w - \eta \nabla_w \loss_{ent}(w;X,Y)
\end{align*}
This gradient ideally requires computation over the entire training set at once; however can be extended to a batch-wise update rule by borrowing key findings from \cite{sgld}. This expectation for the full gradient is computationally intractable, however, Euler discretization of Langevin Stochastic Differential Equation, it can be approximated fairly well as  
\begin{align*}
    w'^{t+1} = w'^t + \eta^t \nabla_{w'} \log p(w'^t) + \sqrt{2\eta} \mathcal{N}(0,\mathbb{I})
\end{align*}
such that after large enough amount of iterations $w^+\to w^\infty$ then $w^\infty \sim p(w')$. One can estimate $\E_{w'\sim p(w')} \left[\gamma (w-w')\right]$ by averaging over many such iterates from this process. This result is stated as it is from \cite{esgd}: $\mathbb{E}_{w'\sim p(w')}[g(w')] = \frac{\sum_t \eta_t g(w'_t)}{\sum_t \eta_t}$.  This leads to the algorithm shown in Algorithm \ref{algo:esgd}. One can further accrue exponentially decaying weighted averaging of $g(w'_t)$ to estimate $\mathbb{E}_{w'\sim p(w')}[g(w')]$. This entire procedure is described in Algorithm \ref{algo:esgd}.

This algorithm is then further guaranteed to find wide minima neighborhoods of $w$ by design, as sketched out by the proofs in \cite{esgd}. 
\subsection{Stochastic Gradient Langevin Dynamics}

Stochastic Gradient Langevin Dynamics combines techniques of Stochastic Gradient  Descent and Langevin Dynamics \cite{sgld}. Given a probability distribution $\pi=p(\theta;X)$, the following update rule allows us to sample from the distribution $\pi$:
\begin{align} \label{eq:ld}
\theta^{t+1} = \theta^{t} +\eta \nabla_{\theta} p(\theta^{t};X) + \sqrt{2\eta'}\varepsilon
\end{align}
where $\eta'$ is step size and $\varepsilon$ is normally distributed. Then, as $t\to \infty$, $\theta \sim \pi$.

While this update rule in itself suffices, if the parameters are conditioned on a a training sample set $X$, which is typically large, the gradient term in Eq. \ref{eq:ld} is expensive to compute. \cite{sgld} shows that the following batch-wise update rule:
$$
\theta^{t+1} = \theta^{t} +\eta \nabla_{\theta} p(\theta^{t};X_{B_j}) + \sqrt{2\eta'}\varepsilon
$$
suffices to produce a good approximation to the samples $\theta^{t\to \infty} \sim \pi$. In Algorithm \ref{algo:desgd}, $\theta$ is the set of perturbed points $X'$. In each internal iteration, we look at subset of trainable parameters $X'_{B_j}$. We update the estimate for $X'_{B_j}$ by only considering data-points $X_{B_j}$ at a time. In the current formulation the set of iterable parameters $X'_{B_j}$ only 'see' a single batch of data $X_{B_j}$ ; a better estimate would require $X'_{B_j}$ to be updated by iteratively over all possible batches $X_{B_k}$, $k=1,2....J$. However in practice we observe that just using the corresponding batch $X_{B_{k=j}}$ suffices. In future work, we will explore the theoretical implications of this algorithmic design.

\subsection{PGD-Adversarial Training} \label{sec:appendixB}

\begin{algorithm}[t]
	\caption{PGD AT}
	\label{algo:pgd-sgd}
	\begin{algorithmic}[1]
		
		\STATE \textbf{Input:} $[X_{B_1},X_{B_2}\dots X_{B_J}],f,\eta,\eta', w=w^0, \epsilon$
		\FOR{$t=0,\cdots T-1$} 
		\FOR{$j=1, \cdots J$}
		\STATE $x'^{0} \gets x$  \COMMENT{$\forall x\in X_{B_j}$}
		\FOR{$k = 0, \cdots,K-1$}
		\STATE	$dx'^k \gets \frac{1}{n}\sum_{i=1}^n \nabla_{x=x^k}L(f(w^t;x)) $
		
		\STATE	$x'^{k+1} \gets x'^k + \eta' dx'^k$ \COMMENT{Gradient ascent}
		
		\STATE project$_{x+\Delta}(x', \epsilon)$
		\ENDFOR
		\STATE $\mu^t \gets L(w^t,x'^{K})$ \COMMENT{batch loss for $X_{B_j}$}
		\STATE $dL^{t} \gets \nabla_w \mu^t$ \COMMENT{gradient of batch loss}
		\STATE $w^{t+1} \gets w^t - \eta dL^t$
		\ENDFOR 
		\ENDFOR 
		\STATE \textbf{Output} $\hat{w} \gets w^T$
		
	\end{algorithmic}
\end{algorithm}

In Algorithm \ref{algo:pgd-sgd} we describe the PGD-AT algorithm. In \cite{madry} authors demonstrate that PGD based-attack is the best possible attack that can be given for a given network and dataset combination. Theoretically, 
\begin{equation}
\bar{x}_{\text{worst}} = \arg\max_{\delta \in \Delta_p} L(f(\hat{w};x+\delta), y)
\label{eq:attack2}
\end{equation}
and if this maximization can be solved tractably, then a network trained with the following min-max formulation is said to be robust:
\begin{equation}
\min_{w}\max_{\delta \in \Delta_p} \loss(f(\hat{w};X+\delta,Y), Y)
\label{eq:attack_train}
\end{equation}
Furthermore they show that first order based gradient approaches, such as SGD, are sufficient to suitably optimize the inner maximization over the perturbed dataset. This can be obtained using the following gradient \textit{ascent} update rule:
$$X' = X' + \eta' \nabla_{X'\in X+\delta} \loss(f(w,X'+\delta;Y),Y)$$
(see also Step 7 of Algorithm \ref{algo:pgd-sgd}). Note that when $\Delta_p = \Delta_2$, this projection rule represents a noise-less version of the update rule in ATENT (see Algorithm \ref{algo:desgd}, line 8).

Iterative Fast Gradient Sign (IFGS) method effectively captures a similar projection based approach which performs an update within an $\ell_\infty$ ball. This update is given by:
$$X' = X' + \eta' \text{sign}(\nabla_{X'} \loss(f(w,X'+\delta;Y),Y))$$
Note that this update rule constructs an adversarial example within $\ell_\infty$-ball, during the training procedure. Meanwhile, given our proposed adversarial example sampling criterion in Assumption \ref{assump:gibbs2}, our update rule is slightly different (see also Eq. \ref{eq:linfupdate}).

\newcommand\scaleFactor{0.6}


   


